\documentclass[preprint,12pt,authoryear]{elsarticle}

\usepackage{amssymb}
\usepackage{amsmath}
\usepackage{subcaption}
\usepackage{booktabs}
\journal{Acta Astronautica}

\begin{document}

\begin{frontmatter}

%% Title, authors and addresses

%% use the tnoteref command within \title for footnotes;
%% use the tnotetext command for theassociated footnote;
%% use the fnref command within \author or \affiliation for footnotes;
%% use the fntext command for theassociated footnote;
%% use the corref command within \author for corresponding author footnotes;
%% use the cortext command for theassociated footnote;
%% use the ead command for the email address,
%% and the form \ead[url] for the home page:
%% \title{Title\tnoteref{label1}}
%% \tnotetext[label1]{}
%% \author{Name\corref{cor1}\fnref{label2}}
%% \ead{email address}
%% \ead[url]{home page}
%% \fntext[label2]{}
%% \cortext[cor1]{}
%% \affiliation{organization={},
%%            addressline={}, 
%%            city={},
%%            postcode={}, 
%%            state={},
%%            country={}}
%% \fntext[label3]{}

\title{Learning Robust Satellite Attitude Dynamics with Physics-Informed Normalising Flow}

\author[1]{Carlo Cena\corref{cor1}}%[orcid=0000-0002-7802-5625]
\ead{carlo.cena@polito.it}

\affiliation[1]{organization={Department of Electronics and Telecommunications, Politecnico di Torino},
                 addressline={Corso Duca degli Abruzzi, 24},
                 postcode={10129},
                 city={Torino}, 
                 country={Italy}}

\author[1]{Mauro Martini}%[orcid=0000-0002-6204-3845]
\ead{mauro.martini@polito.it}

\author[1]{Marcello Chiaberge}%[orcid=0000-0002-1921-0126]
\ead{marcello.chiaberge@polito.it}

\cortext[cor1]{Corresponding author}

\begin{abstract}
Attitude control is a fundamental aspect of spacecraft operations. Model Predictive Control (MPC) has emerged as a powerful strategy for these tasks, relying on accurate models of the system dynamics to optimize control actions over a prediction horizon. In scenarios where physics models are incomplete, difficult to derive, or computationally expensive, machine learning offers a flexible alternative by learning the system behavior directly from data. However, purely data-driven models often struggle with generalization and stability, especially when applied to inputs outside their training domain. To address these limitations, we investigate the benefits of incorporating Physics-Informed Neural Networks (PINNs) into the learning of spacecraft attitude dynamics, comparing their performance with that of purely data-driven approaches. Using a Real-valued Non-Volume Preserving (Real NVP) neural network architecture with a self-attention mechanism, we trained several models on simulated data generated with the Basilisk simulator. Two training strategies were considered: a purely data-driven baseline and a physics-informed variant to improve robustness and stability. Our results demonstrate that the inclusion of physics-based information significantly enhances the performance in terms of the mean relative error with the best architectures found by 27.08\%. These advantages are particularly evident when the learned models are integrated into an MPC framework, where PINN-based models consistently outperform their purely data-driven counterparts in terms of control accuracy and robustness, and achieve improved settling times when compared to traditional MPC approaches, yielding improvements of up to 62\%, when subject to observation noise and RWs friction.
\end{abstract}

%%Graphical abstract
%\begin{graphicalabstract}
%\includegraphics{grabs}
%\end{graphicalabstract}

%% Keywords

\end{frontmatter}

%% Add \usepackage{lineno} before \begin{document} and uncomment 
%% following line to enable line numbers
%% \linenumbers

%% main text
%%

\section{Introduction}
The effectiveness of satellites' attitude control system (ACS) is critical to the overall result of space missions, impacting both their operational efficiency and life horizon. Accurate pointing capabilities are essential for a wide range of mission objectives, including maintaining reliable communication links, gathering precise data from onboard scientific instruments, and ensuring proper thermal regulation (\cite{acs_challenges}). However, designing robust ACS for satellites remains a significant challenge because of the spacecraft's complex and non-linear dynamics, as well as the highly variable conditions of the space environment. Among common attitude control actuators, reaction wheels (RWs) are preferred for their high accuracy and moderately fast maneuvers, providing continuous and smooth control (\cite{rw_ref}). However, they also introduce significant non-linearities, such as frictions and saturation effects. In addition to the complexities introduced by actuators, spacecraft in Earth orbit are continuously subjected to various external disturbance torques, such as gravity gradient, atmospheric drag, and magnetic field torques (\cite{acs_challenges}). These environmental torques increase the angular momentum of the spacecraft, making active control by the ACS essential to maintain the desired orientation. Addressing these complexities has historically relied on two main approaches: model-based methods (\cite{IANNELLI2022401, generic_model_sat_att, manuel, java_riccati}) and, more recently, data-driven techniques (\cite{react, rl_varying_masses, WU20241979, imit_learn_sat_att}). However, both exhibit some limitations. Traditional physics-based models, while foundational and offering a structured understanding of the system behavior, often rely on simplifying assumptions. These simplifications can compromise their accuracy and make them difficult to apply effectively in highly dynamic and uncertain environments (\cite{math11122614}). A notable example of these methods is Model Predictive Control (MPC), which relies on an internal dynamics model, whose fidelity is critical for the stability and robustness of the controller. On the other hand, purely data-driven machine learning (ML) algorithms learn directly from experience without requiring an explicit system model. Although they offer versatility in pattern recognition and scalability, these approaches present significant drawbacks in safety-critical applications. Therefore, improving the spacecraft's dynamics model is a key enabler for more reliable and effective attitude control, and is the central objective of the approach proposed in this work.
In response to the limitations of physics-based models and ML data-driven approaches, Physics-Informed Neural Networks (PINNs) (\cite{pinn1}) have emerged as a significant paradigm shift. PINNs integrate the governing physical laws of a system directly into the neural network's learning process. Traditional PINNs learn to predict the state of the system at a particular time instant, this however does not adapt to control and planning algorithms, which only need to explore the evolution of the actuated system in a limited future time horizon. Therefore alternative approaches have emerged: some merge physical models to AI models trained with data-driven losses (\cite{davide, pinn_sat_state_est}), others add to the data-driven loss an unsupervised physics loss term, which ensures that the physical equations are satisfied at certain points throughout the domain (\cite{pinn_control, pinn_loss_eletr, pinn_loss_elet2}). This embedding of physical knowledge has a regularization effect, which leads PINN to be more data-efficient and robust.

In this work, we explore the application of PINNs to the learning of spacecraft attitude dynamics using a Real-valued Non-Volume Preserving (Real NVP) Neural Network (NN) architecture augmented with a self-attention layer. We compare models trained solely with data-driven loss with those trained with both data-driven and physics-informed losses. All experiments are conducted using high-fidelity simulation data generated with the Basilisk simulator (\cite{basilisk}), ensuring realistic and reproducible evaluation conditions. Our results show that PINNs offer superior performance compared to purely model-free approaches when evaluating the model as a regressor to predict the next state of the satellite's attitude dynamics and exhibit improved stability and robustness when integrated into an MPC framework, highlighting the potential of physics-informed learning in advancing autonomous space systems.

\subsection{Contribution}
The key contributions of our paper are as follows.
\begin{enumerate}
    \item We propose a novel approach for learning spacecraft attitude dynamics using a Real NVP neural network with a self-attention mechanism.
    \item We introduce a physics-informed training loss to boost the generalization and robustness properties of the learned dynamics model, optimizing the data-physics losses ratio using the Lagrangian dual approach.
    \item We systematically compare purely data-driven models with their physics-informed counterparts, highlighting the benefits of incorporating physical information into the training process. Showing improvements between 90.22\% and 27.08\% in terms of mean relative error when predicting 10 time steps in self-loop.
    \item We demonstrate the practical utility of the learned models by embedding them into an MPC framework and evaluating their performance and robustness-to-noise in closed-loop attitude control tasks, significantly limiting the spread of the trajectories when injecting Gaussian noise into the state observations.
    \item We show that the trained models enable improvements of the MPC performance with respect to traditional non-linear and linear dynamics, when the spacecraft is subject to parameters estimation and state observation errors of up to 20\% and 3\% respectively, and RWs friction.
\end{enumerate}

\subsection{Paper Organization}
The paper is organized as follows. Section \ref{related_works} gives an overview of the related works. Section \ref{methodology} provides a detailed explanation of the proposed model with the loss functions used. Section \ref{exp_setup} describes the dataset and the metrics used to evaluate the framework. In Section \ref{results}, we present a comprehensive discussion of the results obtained. Finally, Section \ref{conclusion} offers concluding remarks on the outcomes and implications of applying our framework.

\section{Related Works}
\label{related_works}
This section reviews relevant works that used machine learning, both purely data-driven and with physics-informed approaches, and the integration of neural networks with MPC, for control, state estimation, and trajectory optimization, with a focus on the aerospace field.

Neural networks have been successfully applied to various aspects of control and state estimation and are being increasingly adopted for critical tasks. %\cite{dec_transformer} tried to improve reinforcement learning by reframing it as a sequence modeling problem, where optimal actions are predicted based on a sequence of states, actions, and rewards. In the aerospace context, neural networks are being increasingly adopted for critical tasks.
(\cite{pinn_loss_eletr}) uses a physics-informed loss to improve resilience against disturbances when estimating the state of a power-grid with a neural network, showing significant improvements in estimation accuracy and robustness under challenging conditions like three-phase faults and data manipulation attacks. (\cite{pinn_loss_elet2}) introduced Physics-informed Electromagnetic Field Network (PEFNet), a neural network trained to address the path loss estimation problem by using the computational electromagnetic principles in a physics-informed loss. For what concerns the aerospace sector, in (\cite{pinn_rizzo}) a physics-informed neural network was trained to model the non-linear dynamics of quadrotors, embedding conservation laws to enhance generalization and interpretability, outperforming both traditional models and black-box neural networks. (\cite{IZZO2023510}) showed the use of neural representations for time-optimal, constant acceleration rendezvous, highlighting the ability of neural networks to learn complex, non-linear control policies. For what concerns state estimation, (\cite{pinn_sat_state_est}) investigated the application of PINNs for satellite state estimation during continuous thrust maneuvers, showing how the incorporation of physical laws into neural networks can enhance the accuracy and robustness of estimations in orbital mechanics. Unlike them, we apply the physics-informed approach for attitude dynamics and aim to directly learn the state transition function through a physics-informed loss, instead of a perturbation. Furthermore, (\cite{imit_learn_sat_att}) explored imitation learning and generative adversarial NNs for satellite attitude control under unknown perturbations using the physical simulator MuJoCo (\cite{mujoco}), suggesting a promising avenue for robust and adaptive control in uncertain space environments. Finally, (\cite{WU20241979}) uses a fully-connected deep neural network trained with a data-driven approach to estimate the inertia matrix of the final system in the context of docking and berthing.

Concerning the integration of artificial intelligence approaches in MPC in the aerospace field, (\cite{drones7010004}) studied an MPC framework for aerial robots that combines an offline physics-derived model with an online machine learning correction using adaptive sparse identification. (\cite{CHEN2022109947}) proposed a MPC framework that utilizes a neural network to replace the non-linear dynamics of an MPC and primal active sets to efficiently handle complex constraints and dynamics. Extending this, (\cite{pinn_multilink}) explored PINN-based MPC for multi-link manipulators, demonstrating the potential of combining data-driven and physics-informed models for improved control performance. A more recent advancement in this area is presented by (\cite{davide}), who introduced a Transformer-Based MPC approach. They used the sequence modeling capabilities of transformers to generate better initial solutions to be used as starting trajectories in a MPC, and learned the terminal cost, improving runtime and convergence. Though they used a purely data-driven loss to train the neural network. These works highlight a growing trend towards using advanced machine learning techniques, particularly transformer architectures and physics-informed approaches, to overcome challenges in traditional control and estimation problems.

\section{Methodology}
\label{methodology}
In this section, we provide a detailed explanation of the proposed framework for physics-informed attitude dynamics learning. First, we formally frame the problem of satellite attitude dynamics. Then, the neural network model adopted is briefly described, and the physics-biased loss function used in the training is defined. Finally, the non-linear Model Predictive Control framework integrated with the learned dynamics is presented.

\subsection{Satellite Attitude Dynamics}
\label{sec:dynamics_back}
Here the necessary background on the attitude dynamics of a rigid spacecraft in Earth orbit is provided. 
A spacecraft is subject to both actuators control torques and external torques. The environmental torques acting on the satellite as disturbances can have different sources depending on its position and velocity: gravity gradient, atmospheric drag, and magnetic field torques. We are not providing a detailed formulation of these forces, though they are present in the simulator used to collect trajectories for training and testing, and they have an effect on the learned dynamics.

Given the actuators control torque in the body coordinate frame $N_{c}$, the satellite inertia matrix $I_{s}$, the reaction wheels inertia matrix $I_{rw}$, the angular velocity of the satellite $\omega$ and of the reaction wheels $\omega_{rw}$, the total external torque $N_{e}$, and defining the skew-symmetric matrix $S(\omega)$ in eq. (\ref{eq:skewsym}) we have that the dynamics of the satellite actuated by the Reaction Wheels (RWs) is shown in eq. (\ref{eq:self_sup}).

\begin{equation}
    \label{eq:skewsym}
    S(\omega) = \begin{bmatrix} 
                        0 & -w_3 & w_2\\
                        w_3 & 0 & -w_1\\
                        -w_2 & w_1 & 0
                        \end{bmatrix}
\end{equation}

\begin{equation}
\label{eq:self_sup}
    \dot \omega = -I_{s}^{-1}S(\omega)I_{s}\omega - I_{s}^{-1}S(\omega)I_{rw}\omega_{rw}+I_{s}^{-1}N_{c}+I_{s}^{-1}N_{e}
\end{equation}

Given the RWs torque $u_{rw}$ it is possible to calculate the torque acting on the spacecraft as shown in eq. (\ref{eq:nc}).
\begin{equation}
\label{eq:nc}
N_{c} =  -u_{rw} + I_{rw}\dot \omega
\end{equation}

The dynamics of the reaction wheels angular velocity is given in eq. (\ref{eq:wrw}).

\begin{equation}
\label{eq:wrw}
\dot \omega_{rw} = I_{rw}^{-1}u_{rw} - \dot \omega
\end{equation}

It is possible to obtain the angular velocity of the spacecraft and of the reaction wheels by integrating over time eq. (\ref{eq:self_sup}) and (\ref{eq:wrw}). This allows to compute the attitude of the spacecraft, defined by quaternion \textit{q}, by integrating eq. (\ref{eq:quat}).

\begin{equation}
    \label{eq:quat}
    \dot q = \frac{1}{2} \Omega[\omega]q
\end{equation}

Where $\Omega[\omega]$ is given by eq. (\ref{eq:omegaw}).
\begin{equation}
    \label{eq:omegaw}
    \Omega[\omega] = \begin{bmatrix} 
                        0 & -\omega_0 & -\omega_1 & -\omega_2\\
                        \omega_0 & 0 & \omega_2 & -\omega_1\\
                        \omega_1 & -\omega_2 & 0 & \omega_0\\
                        \omega_2 & \omega_1 & -\omega_0 & 0
                        \end{bmatrix}
\end{equation}

Finally, the total angular momentum of the spacecraft \textit{h} is composed of the angular momentum of the spacecraft and the angular momentum of the reaction wheels as shown in eq. (\ref{eq:ang_mom}).

\begin{equation}
    \label{eq:ang_mom}
    h = I_{s} \omega + I_{rw} \omega_{rw}
\end{equation}

\subsection{Problem Formulation}

\begin{figure*}
    \centering
    \includegraphics[width=0.9\linewidth]{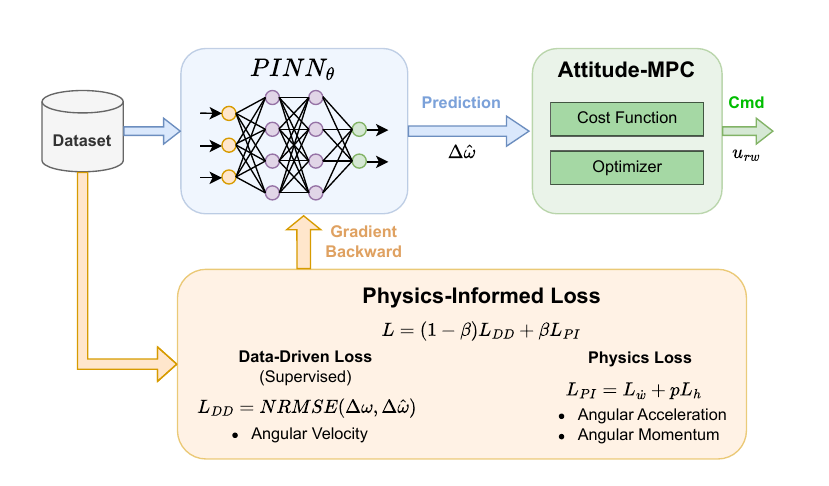}
    \caption{A schematic description of the proposed PINN for satellite dynamics learning and attitude control. The PINN models, trained using a weighted sum of the data-driven and physics-informed losses, is used as dynamics model in a non-linear MPC.}
    \label{fig:PINN_scheme}
\end{figure*}

Developing a reliable attitude control system for satellites remains exceptionally challenging given the spacecraft’s inherently non-linear and intricate dynamic behavior, coupled with the unpredictable conditions of the space environment (\cite{acs_challenges}). 
Enhancing the accuracy of the spacecraft dynamics model is therefore indispensable for achieving more precise attitude regulation, also adopting advanced model-based control techniques such as model predictive control (MPC). This becomes particularly important when the spacecraft state and parameters are subject to errors and estimation inaccuracies.

In this work, we propose a novel physics-informed neural network (PINN) approach to approximate the full dynamics model of the spacecraft. Besides the typical data-driven loss, a physics-based loss is designed to act as an inductive bias during the training process of the model, leading to enhanced generalization and robustness to noise compared to vanilla data-driven formulations. Figure \ref{fig:PINN_scheme} describes the complete pipeline of the learning approach. The PINN receives the spacecraft state, comprising angular velocities of both satellite and reaction wheels, together with the angular acceleration of the satellite and the RWs control torque as input. It predicts the resulting changes in angular velocities of the satellite, to be directly integrated in a non-linear MPC framework usually adopted in advanced attitude control solutions.

\subsection{Neural Network Architecture}
\label{sec:nn_architecture}
A neural network model is used to learn the attitude dynamics of the spacecraft. We experimented with the Real NVP architecture and with self-attention layers, with the aim of limiting the computational cost of the inference by adopting an efficient architecture.

Real NVP belongs to the family of normalizing flow (\cite{realnvp_paper}), aimed at modeling high-dimensional data distributions through a sequence of bijective transformations. This type of model facilitates both density estimation and sample generation using affine coupling layers, which operate by partitioning the input variables: one subset undergoes transformation via a scale and a translation neural networks, while the other remains unchanged. Both networks receive half of the input and apply their respective transformations to the complementary half, ensuring that the overall mapping remains invertible and computationally efficient.

\begin{figure*}[!htb]
    \centering
    \centerline{\includegraphics[width=\linewidth]{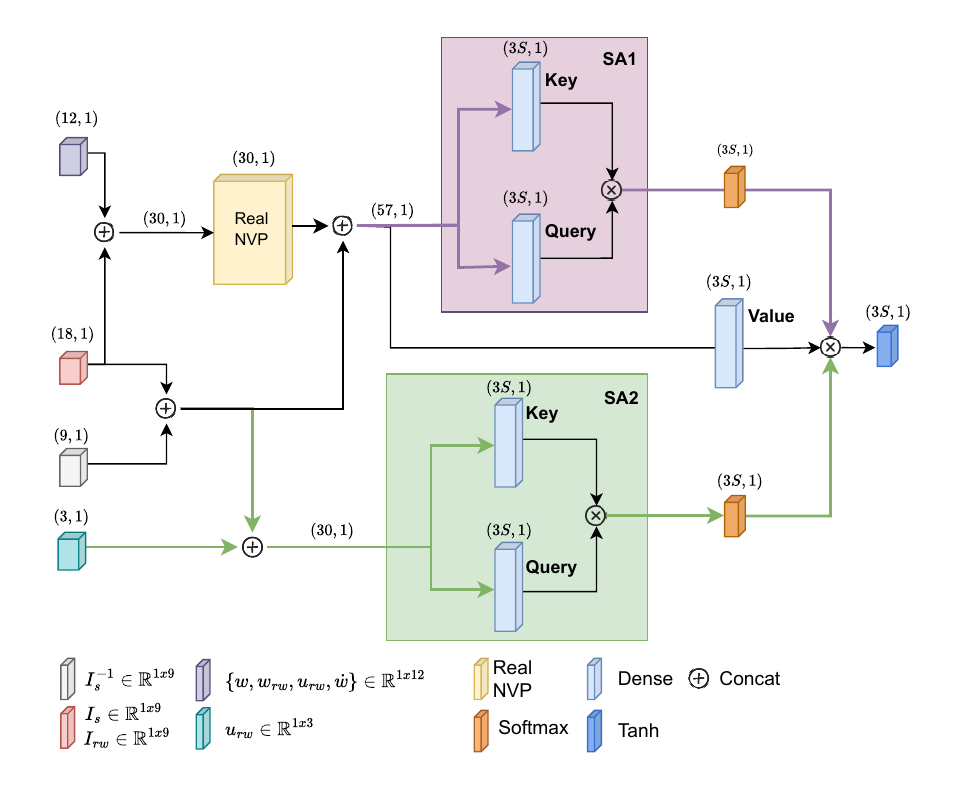}}
        \caption{Neural Network architecture diagrams. NVPSA1 (violet flow) uses the output of the Real NVP model to weigh the delta state predicted through a self-attention approach. NVPSA2 (green flow) instead uses the input torque and the inertia matrices of the spacecraft and of the RWs as input of the Query and Key layers. \textit{S} is the length of the predicted trajectory, in our case \(S=10\).}
	\label{fig:model_attention}
\end{figure*}

Self-Attention (\cite{selfattention2, selfattention1}) is a mechanism that allows a model to weigh the importance of different elements within an input sequence when encoding contextual information. It captures long-range dependencies and enables dynamic representation learning by computing pairwise interactions between all tokens. In our framework, it is used to let the network focus on correlated outputs and scale correctly the prediction when dealing with low intensity input signals. We demonstrate the benefits of this architecture in Section \ref{results}, showing that the model is more precise in predicting null changes. We implemented and experimented with two slightly different versions by changing the input of the Key and Query layers, as can be seen by looking at the differences between SA1 and SA2 in Fig. \ref{fig:model_attention}.
In particular, in model \textit{NVPSA1}, which uses the flow SA1, we use the output of the Real NVP model as input to compute Value, Query, and Key. While in \textit{NVPSA2}, which uses the flow SA2, the FC layer for the Query and Key received as input the concatenation of the commanded torques, the inertia matrix of the satellite and of the RWs, and the inverse of the satellite inertia matrix. 
At time step \textit{t} all models aimed at predicting the resulting change in angular velocity $\Delta{\hat{\omega}_{t+1}}$ using as input a state composed of the current satellite angular velocity $\omega_{t}$, RWs velocity $\omega_{rw,t}$, the RWs commanded torque $u_{rw,t}$, and the angular acceleration of the spacecraft $\dot{\omega}_{t}$, estimated by computing the first-order backward difference.

\subsection{Physics-Informed Training}
\label{sec:pinn_training}
As previously stated, we combined two loss functions. A classic data-driven loss, \(L_{DD}\), shown in eq. (\ref{eq:data_driven}) is defined as the Normalized Root Mean Squared Error  (NRMSE) of the predicted change in angular velocity $\Delta \hat \omega$ when compared to ground truth data obtained from the dataset. We used the standard deviation of the ground truth change in angular velocity $\sigma_{\Delta \omega}$ to normalize the Root Mean Squared Error (RMSE).

\begin{equation}
    L_{\text{DD}} = \frac{\sqrt{\frac{1}{B}\sum_{i}^{B}(\Delta \hat \omega_{i} - \Delta \omega_{i})^2}}{\sigma_{\Delta \omega}}
\label{eq:data_driven}
\end{equation}

A physics-informed penalty term \(L_{PI}\) is designed to regularize the training of the dynamics model and avoid overfitting by embedding the physical laws of the system directly in the loss function, shown in eq. (\ref{eq:phys_inf_loss}). It has been derived from the satellite attitude dynamics discussed in Section \ref{sec:dynamics_back}, and it is composed of a weighted sum of two components: (I) the NRMSE of the predicted change in angular acceleration $L_{\dot \omega}$, eq. (\ref{eq:dotw}), where $\dot \omega$ is calculated using eq. (\ref{eq:self_sup}); (II) the MSE of the change in angular momentum $L_{h}$, eq. (\ref{eq:totangmom}). 

\begin{equation}
\label{eq:phys_inf_loss}
L_{PI} = L_{\dot \omega} + pL_{h}
\end{equation}

The MSE of the change in angular momentum is weighted by a factor \textit{p} equal to $1e^{-2}$, found through a grid search. To limit the training operations, the total external torque $N_{e}$ in eq. (\ref{eq:self_sup}) is set to 0, although external disturbances were present in the environment, and their effect was present in the training data. We assume that the relative weight of the two losses will let the model learn their effect through the data-driven loss, to this end, we experimented by giving a higher weight to $L_{DD}$ rather than to $L_{PI}$.
\begin{equation}
\label{eq:dotw}
L_{\dot \omega} = \frac{\sqrt{\frac{1}{B}\sum_{i}^{B}(\hat{\dot \omega} - \dot \omega)^2}}{\sigma_{\dot \omega}}
\end{equation}

\begin{equation}
\label{eq:totangmom}
L_{h} = \frac{1}{B}\sum_{i}^{B}(||I_{s} \cdot (\omega+\Delta \hat \omega) + I_{rw} \cdot \hat \omega_{rw}|| - ||I_{s} \cdot (\omega+\Delta \omega) + I_{rw} \cdot \omega_{rw}||)^2
\end{equation}

In equations (\ref{eq:dotw}) and (\ref{eq:totangmom}), $\Delta \hat \omega$ and $\hat{\dot \omega}$ represent the predicted change in angular velocity and the predicted angular acceleration, calculated as $\hat{\dot \omega} = \Delta \hat \omega / \Delta t$. $\Delta \hat \omega_{rw}$ is the change in angular velocity of the RWs computed with eq. (\ref{eq:wrw}) using $\hat{\dot \omega}$. \textit{B} is the batch size.

Equation (\ref{eq:total_loss}) shows the total loss used in our physics-informed experiments, as the weighted sum of the data-driven and physics-informed losses.

\begin{equation}
    \label{eq:total_loss}
    L = \alpha L_{DD} + \beta L_{PI}
\end{equation}

%Here $\beta$ is a parameter automatically updated during training with the Lagrangian dual approach \cite{lagr_paper}, which learns the best multipliers with a sub-gradient method.
Here $\alpha$ and $\beta$ are used to change the relative weight of the two loss functions. We experimented with fixed hand-selected values and with the Lagrangian dual approach (\cite{lagr_paper}), which automatically learns the best multipliers with a sub-gradient method during training, dynamically changing the relative importance of the two losses based on the performance of the model. In the latter case $\alpha = (1-\beta)$, and $\beta$ was constrained in the interval [0, 1].

\subsection{Non-linear MPC with Learned Dynamics} \label{mpc_cost_def}
\begin{figure*}[!htb]
    \centering
    \centerline{\includegraphics[width=\linewidth]{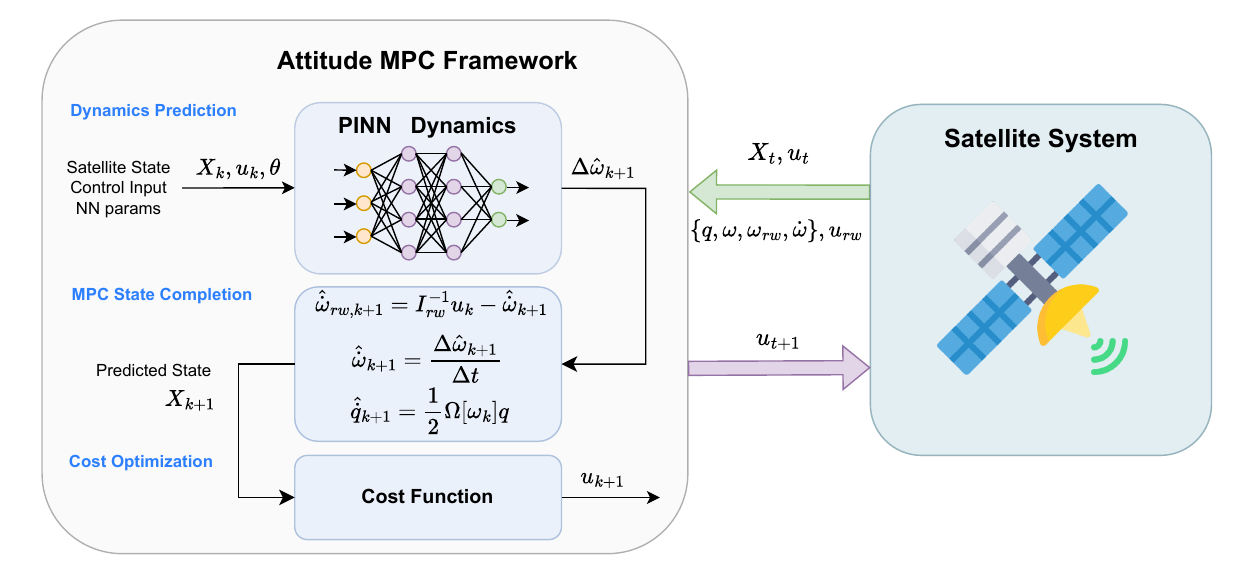}}
        \caption{Attitude MPC with learned dynamics, a schematic of the complete framework.}
	\label{fig:MPC_scheme}
\end{figure*}

The main focus of this work is learning the attitude dynamics of a satellite through a deep neural network, which can provide substantial benefits to diverse spacecraft control operations. We demonstrate evaluating the best models obtained in a rest-to-rest spacecraft maneuvers application, coupling the learned dynamics with a MPC framework, Fig. \ref{fig:MPC_scheme}. 
This concrete application has also been carried out with the aim of testing the effect of the learned dynamics in a model-based controller and evaluating its robustness to state estimation errors and model uncertainties. Section \ref{exp_setup} will describe the metrics used in the two parts of the project in more detail.
 
The state used by the MPC is composed of the current quaternion \textit{q}, spacecraft angular velocity $\omega$, RWs angular velocity $\omega_{rw}$, and satellite angular acceleration $\dot \omega$ as $x = \{q, \omega, \omega_{rw}, \dot \omega\}$. The PINN is used to estimate the dynamics of the spacecraft and of the RWs starting from the current state $x_{t}$ and the current RWs torque $u_{t}$, which is then integrated to obtained the new state $x_{t+1}$, as shown in equations \ref{eq:dw_mpc}-\ref{eq:dotq_mpc}.

\begin{align}
    \label{eq:dw_mpc}
    %\hat{X}_{t+1} = PINN_{\theta}(\hat{X}_t, u_t,\theta)
    % &\Delta \hat{\omega}_{t+1} = PINN_{\theta}(X_t, u_t,\theta) \\
    % &\hat{\dot{\omega}}_{t+1} = \frac{\Delta \hat{\omega}_{t+1}}{\Delta t} \\
    % &\hat \omega_{t+1} = \omega_{t} + \Delta \hat{\omega}_{t+1}\\
    % &\hat \omega_{rw} = \omega_{rw} + \hat{\dot{\omega}}_{rw} \cdot \Delta t\\
    % &\hat q_{t+1} = q_{t} + \hat{\dot{q}}_{t+1} \cdot \Delta t\\
    % &\\hat{\dot{X}}_{t+1} = \{\hat q_{t+1}, \hat \omega_{t+1}, \hat \omega_{rw}, \hat{\dot \omega_{t+1}}\}
    &\Delta \hat{\omega}_{t+1} = PINN_{\theta}(\omega_t, \omega_{rw, t}, \dot \omega_t\, u_t,\theta) \\
    \label{eq:dotw_mpc}
    &\hat{\dot{\omega}}_{t+1} = \frac{\Delta \hat{\omega}_{t+1}}{\Delta t} \\
    \label{eq:dotwrw_mpc}
    %&\hat{\dot{\omega}}_{rw} \text{ from eq. \ref{eq:wrw}}\\
    %&\hat{\dot{q}}_{t+1} \text{ from eq. \ref{eq:quat}} \\
    &\hat{\dot{\omega}}_{rw,t+1} = I_{rw}^{-1} u_{t} - \hat{\dot{\omega}}_{t+1} \\
    \label{eq:dotq_mpc}
    &\hat{\dot{q}}_{t+1} = \frac{1}{2} \Omega[\omega_{t}]q
\end{align}

In particular, our PINN is used to estimate the change in angular velocity $\Delta \hat \omega_{t+1}$, which is subsequently used to compute the angular acceleration $\hat{\dot{\omega}}_{t+1}$ (eq. \ref{eq:dotw_mpc}), and the RWs accelerations $\hat{\dot{\omega}}_{rw}$, by substituting the estimated angular acceleration in eq. (\ref{eq:wrw}). The quaternion dynamics $\hat{\dot{q}}_{t+1}$ is calculated using the angular velocity received in input, thus it depends directly on the predicted output of the PINN starting from the second step of the MPC horizon.

The MPC's cost function is defined in eq. (\ref{eq:mpc_cost_f}).
\begin{equation}
\label{eq:mpc_cost_f}
C = \sum_{k=0}^{n-1}\left( x_{k}^{T}Qx_{k} + u_{k}^TCu_{k}
    + \Delta u_k^T R \Delta u_k\right)
    + x_{n}^{T}Qx_{n}
\end{equation}
Where $u_k$ is an abbreviation of the commanded RWs torque $u_{rw}$ at time step \textit{k}, and $\Delta u_{k}$ is the difference between the RWs torque required at the previous time step and the current one. \textit{C} and \textit{R} are diagonal cost matrices used to reduce the torques $u_{rw}$ and to limit their variation, smoothing their trajectories. Their non-zero elements are equal to $1e-1$, while $Q^{13 \times 13} =$ \text{diag}(10000, 10000, 10000, 10000, $1e^{-2}, 1e^{-2}, 1e^{-2}, 1e^{-4}, 1e^{-4}, 1e^{-4}, 1e^{-2}, 1e^{-2}, 1e^{-2}$). Finally, \textit{n} is the horizon of the MPC, which in our experiments is set to $n = 10$.

\section{Experimental setups}
\label{exp_setup}
In this section we describe in depth the dataset used, the NNs architectures, the traditional MPC implementations, the experiments and the metrics computed to evaluate the results.

\subsection{Dataset}
To train and test the models, we generated a dataset using the Basilisk simulator (\cite{basilisk}). The simulation involved a 58kg satellite in Low Earth Orbit actuated only by RWs which performed attitude maneuvers controlled by a MRP feedback control module available in Basilisk. In addition to the commanded torques, on the satellite acted several environmental disturbances, such as the gravity gradient, magnetic disturbances, and the atmospheric drag. The relevant satellite parameters are shown in Table \ref{tab:datasets_info}.
\begin{table}[htb]
    \caption{We show the satellite inertia matrix \(I_S\), the RWs inertia matrix \(I_{rw}\), the satellite mass, the maximum torque acting on the RWs \(u_{rw}\), the maximum RWs speed \(\omega_{rw}\), and the controller timestep \(\Delta t\).}
    \centering
    \begin{tabular}{@{}cccccc@{}}
        \toprule
        \(I_{s}\) & \(I_{rw}\) & \begin{tabular}{@{}c@{}}Mass\\\ [Kg]\end{tabular} & \begin{tabular}{@{}c@{}}Max \(u_{rw}\)\\\ [Nm]\end{tabular} & \begin{tabular}{@{}c@{}}Max \(\omega_{rw}\)\\\ [rpm]\end{tabular} & \begin{tabular}{@{}c@{}}\(\Delta t\)\\\ [s]\end{tabular}\\
        \midrule\midrule
        \begin{tabular}{@{}c@{}}5.700, 0.045, 0.002\\0.045, 3.300, 0.012\\0.002, 0.012, 6.100\end{tabular} & \begin{tabular}{@{}c@{}}0.001, 0., 0.\\0., 0.001, 0.\\0., 0., 0.001\end{tabular} & 58 & 0.05 & 6000 & 0.1\\
        \bottomrule
    \end{tabular}
    \label{tab:datasets_info}
\end{table}
The rationale for this dataset design is to reproduce a representative operational environment for small satellites performing attitude control solely with RWs. By including both nominal and perturbed configurations, the dataset allows assessing not only nominal performance but also the robustness of the learned models to model uncertainties. We ran 300 runs randomly varying the initial attitude, and orbital position, to generate an initial dataset, and other 50 runs with a different satellite inertia matrix and mass, to test the robustness of the trained models to errors of approximately 10\% on all axes, as these parameters are typically not known with absolute precision and may change during the operational life. These trajectories cover a broad portion of the attitude space and reaction wheel state space, providing a diverse training and testing distribution of control inputs and system responses. In each simulation, the satellite started with a random angular velocity of the reaction wheels, sampled uniformly from [-300, 300] rotations per minute. Each simulation lasted 3 minutes with a sampling and control time of 0.1 seconds, and a simulation time step of 0.001 seconds with Runge-Kutta 4 as integrator.

We construct the dataset by creating input-output pairs as follows: the input $x_{i}$ has shape (1, 12) and is defined as $x_i = \{\omega, \omega_{rw}, u_{rw}, \dot \omega\}$, while the target $y_i$ is equal to the change in satellite angular velocity, i.e. $y_i = \Delta \omega_{i}$. The training dataset was split 67-33\% respectively for training and validation.

\subsection{Models}
In our experiments, we tested several models with varying architectures. The objectives were to obtain good performances in estimating the next attitude state given the current one and the commanded torque, and to maintain a low inference time. We experimented with a Real NVP whose scale and translation networks are composed only of Fully-Connected (FC) layers, and two combinations of Real NVP with a self-attention mechanism to decouple the predicted outputs. This architecture was selected as initial experiments showed that the Real NVP led to an improvement of 26.43\% in terms of MRE and 30.59\% in terms of physics-informed loss, when compared against a MLP, on the test set for the best pair of architectures found by a grid search. Notably, for both models, the identified architecture was considerably smaller than the maximum permitted by the defined search space. Moreover, the Real NVP remains computationally efficient and fast, making it a practical choice for deployment despite its enhanced expressiveness.

In addition to the number of FC and coupling layers, and units in each FC layer, we experimented with the introduction of the satellite and RWs inertia information into the network. In particular, we concatenated it to the input of the model and before the final layer. In the baseline, without self-attention mechanism, a skip connection was used to concatenate the input to the output of the Real NVP before the final FC layer. The best configuration found, shown in Figure \ref{fig:model_attention}, concatenates \(I_{s}\) and \(I_{rw}\) with the input of the model and \(I_{s}\), \(I_{rw}\) and \(I_{s}^{-1}\) to the features generated by the model before the last layer.

Finally, with the best models found, we experimented with the number of predicted steps, \(S\) in Figure \ref{fig:model_attention}. The model was trained to predict with a single inference all future \(S\) steps, and the two losses were computed by iteratively propagating the state while keeping the torques constant. The best results were obtained with \(S=10\). During inference only the first predicted state was used to evaluate the model or as estimated state for the MPC.

\subsection{Traditional MPC} \label{sec:trad_mpc}
In this work we compare our framework to two different MPC without AI, one uses a state space form (linear MPC) and another one uses a non-linear approach. Given the MPC context introduced in Section \ref{mpc_cost_def}, we use the same input state and cost matrices with the non-linear dynamics, implemented following Section \ref{sec:dynamics_back}, and change them when using the state space form. In particular, we used as input state \(x = \{q, w\}\) and  all terms of matrix $Q^{6 \times 6}$ were set to 1000, all other parameters were left unchanged. To implement the discrete state space model, given in eq. \ref{eq:state_space_model}, we followed (\cite{manuel}).
\begin{equation}
\label{eq:state_space_model}
    \dot x_{t+1} = A_{d}x_t + B_{d}u_t
\end{equation}
In eq. \ref{eq:state_space_model} matrix $A_d$ and matrix $B_d$ are computed from eq. \ref{eq:A_cont} and eq. \ref{eq:B_cont} using the zero-order hold control implementation, i.e. keeping the control torque constant according to the control frequency, as shown in eq. \ref{eq:zoh}. Given $w_s$, which is the scalar angular velocity of the spacecraft about the Earth's center, and \textit{I}, which is the satellite inertia matrix $I_s$.
\begin{equation}
\label{eq:A_cont}
    A = \begin{bmatrix} 
        \begin{array}{cc}0^{3 \times 3} & 0.5 I^{3 \times 3}\end{array}  \\
        M^{3 \times 6}\\
        \end{bmatrix}
\end{equation}
\begin{equation}
\label{eq:B_cont}
B = \begin{bmatrix} 
    0^{3 \times 3} \\
    \text{diag}\left(\frac{1}{I_0}, \frac{1}{I_1}, \frac{1}{I_2}\right)
        \end{bmatrix}
\end{equation}
where \textit{M} is defined in eq. \ref{eq:m}.
\begin{equation}
\label{eq:m}
M = \begin{bmatrix}
 -8w_s^2\frac{I_1-I_2}{I_0} & 0 & 0 & 0 & 0 & w_s\frac{I_0+I_2-I_1}{I_0}\\
        0 & -6w_s^2\frac{I_0-I_2}{I_1} & 0 & 0 & 0 & 0\\
        0 & 0 & -2w_s^2\frac{I_1-I_0}{I_2} & -w_s\frac{I_2+I_0-I_1}{I_2} & 0 & 0
\end{bmatrix}
\end{equation}
\begin{equation}
\label{eq:zoh}
    A_d = e^{A\delta t}, B_d = \int_0^{\delta t} e^{A\sigma}d \sigma
\end{equation}

\subsection{Evaluation Metrics \& Experiments} \label{sec:eval_metrics_exp}
To evaluate our models, we used a combination of metrics to evaluate their performance both as standalone regressors and to compare the performances of the resulting MPC controllers. These two evaluations represent the two steps that we followed during our experiments.

Firstly, we trained several neural networks with varying architectures and hyper-parameters to find the best performing framework to estimate the next attitude, given the current one and the commanded RWs torque. We will refer to this step as Dynamics Learning Experiments. The metrics used for this step are the Mean Relative Error (MRE) and the physics error, i.e. the physics-informed loss. The results were averaged over all time steps and simulations. For both metrics we will show both the results on the single-step prediction and the results when predicting the next 10 steps under the assumption of constant command torque. This was done because preliminary results showed that the models were unable to learn from the small changes associated with a single-step prediction. This will also be shown in Section \ref{results}.

Secondly, in the experiments named MPC Experiments, we used the best performing neural networks to test various non-linear MPCs evaluated with the Stability Performance Error, which is the difference between the instantaneous performance error at a given time \textit{t} and the error value at an earlier time $t-\Delta t$, as defined in "ECSS-E-HB-60-10A" (\cite{ecss_10a}). This has been computed using $\Delta t = 10$. The performance error is defined as the difference between the desired state and the current one.

We performed 300 Monte Carlo simulations for each model to test their robustness-to-noise, adding a random error of up to 3\% from a Gaussian distribution to each state variable, and a continuously distributed random error of up to 10\% to the inertia matrix, and up to 20\% to the satellite mass. Moreover, friction was inserted in the RWs following a linear dynamics up to $50\%\pm12,5\%$ of the RWs maximum velocity. The initial attitude was varied randomly in the range ($\frac{\pi}{8}$, $\frac{\pi}{2}$). It should be noted that the neural networks were trained with data generated in simulations without noise and RWs friction. To evaluate these experiments, we will show the trajectories distribution and the stability performance error, along with the average steady-state error over the last minute of simulation and the time required to converge to a attitude error below 1 degree (Settling Time).

\subsection{Implementation Details}
\label{Impl_det}
The experiments were performed on a computer with an Intel i7-9700K 3.60GHz CPU, 64 GB of RAM, and a GeForce RTX 2080 Ti. The models and loss functions have been implemented in torch and we used the "do-mpc" library to implement the MPC with AI as dynamics estimator. The models have been trained on the GPU using a batch size of 16384, found during the hyper-parameter search, and tested on CPU with an inference time of \(0.66 \pm 0.03\)ms for the biggest models, i.e. the Real NVP with self-attention mechanism. Finally, the models were tested on relevant edge devices to acquire data about their computational cost. These results are shown in Section \ref{inf_time}. 

\section{Experiments and Results}
\label{results}
Here we show and discuss the results obtained in the two groups of experiments. Each section presents the results obtained in the relative experiments using the metrics described in Section \ref{exp_setup}.

\subsection{Dynamics Learning Experiments}
We performed several experiments to identify the best architecture, and the best performing model found has 4 coupling layers, whose scale and translation network have 2 FC layers with 64 units each. This model was used to study the impact of concatenating the inertia parameters of the satellite and the RWs of the training set, respectively \(I_{s}\) and \(I_{rw}\), to the input state, and the use of different pairs of weights for the data-driven and physics-informed loss, along with the use of the Lagrangian dual method (\cite{lagr_paper}). The results showed that in all cases the introduction of the inertia parameters led to improved results. It should be noted that this addition helped the models even when the \(I_{s}\) parameters were subject to an error of approximately 10\%, that is, the test set.

\begin{figure}[!h]
    \centering
    \begin{subfigure}[t]{0.48\textwidth}
        \centering
        \includegraphics[width=1\textwidth]{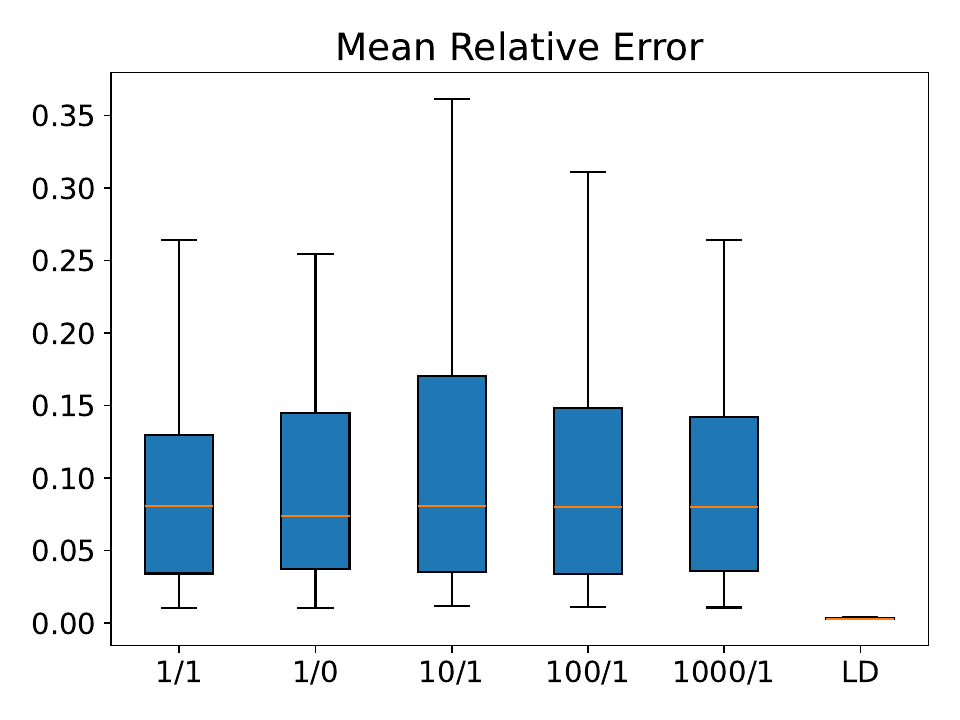}
        \captionsetup{font=footnotesize, justification=centering}
    \end{subfigure}
    \begin{subfigure}[t]{0.48\textwidth}
        \centering
        \includegraphics[width=1\textwidth]{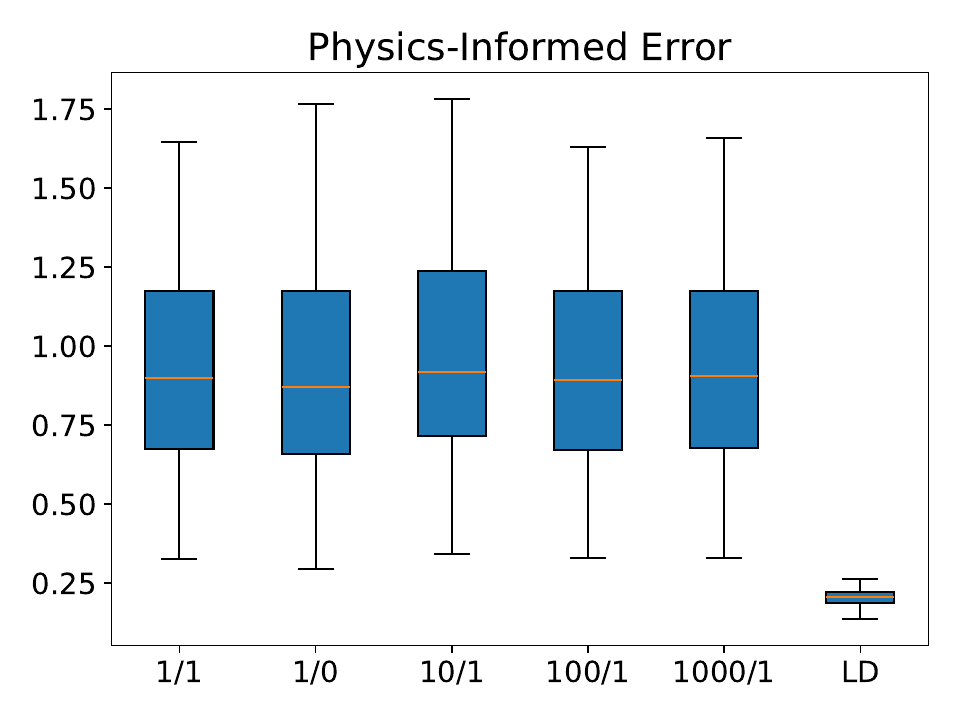}
        \captionsetup{font=footnotesize, justification=centering}
    \end{subfigure}
    \begin{subfigure}[t]{0.48\textwidth}
        \centering
        \includegraphics[width=1\textwidth]{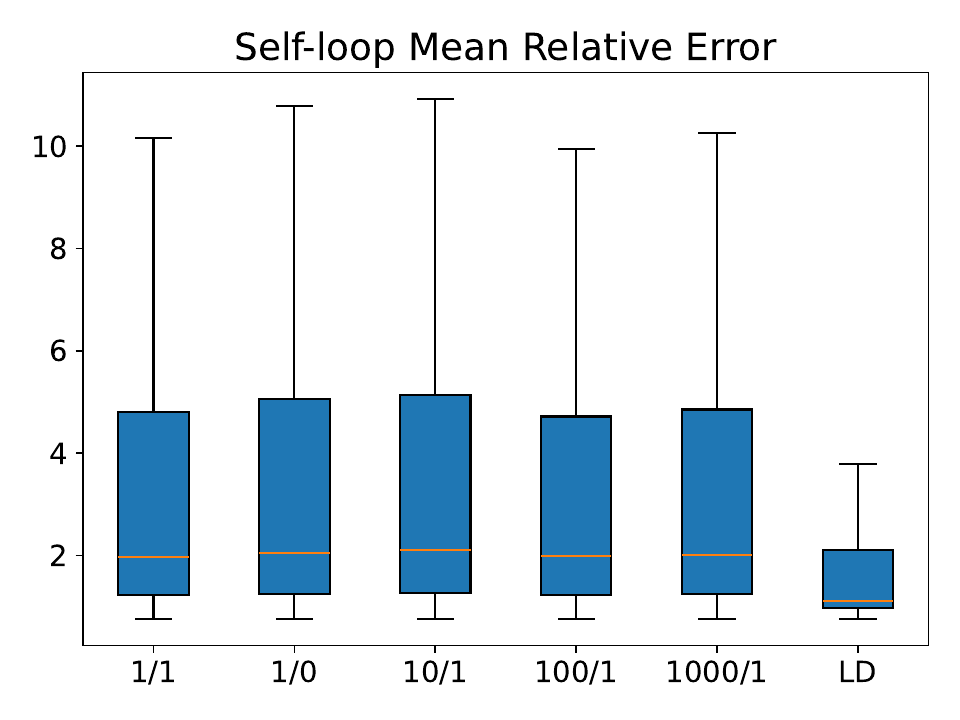}
        \captionsetup{font=footnotesize, justification=centering}
    \end{subfigure}
    \begin{subfigure}[t]{0.48\textwidth}
        \centering
        \includegraphics[width=1\textwidth]{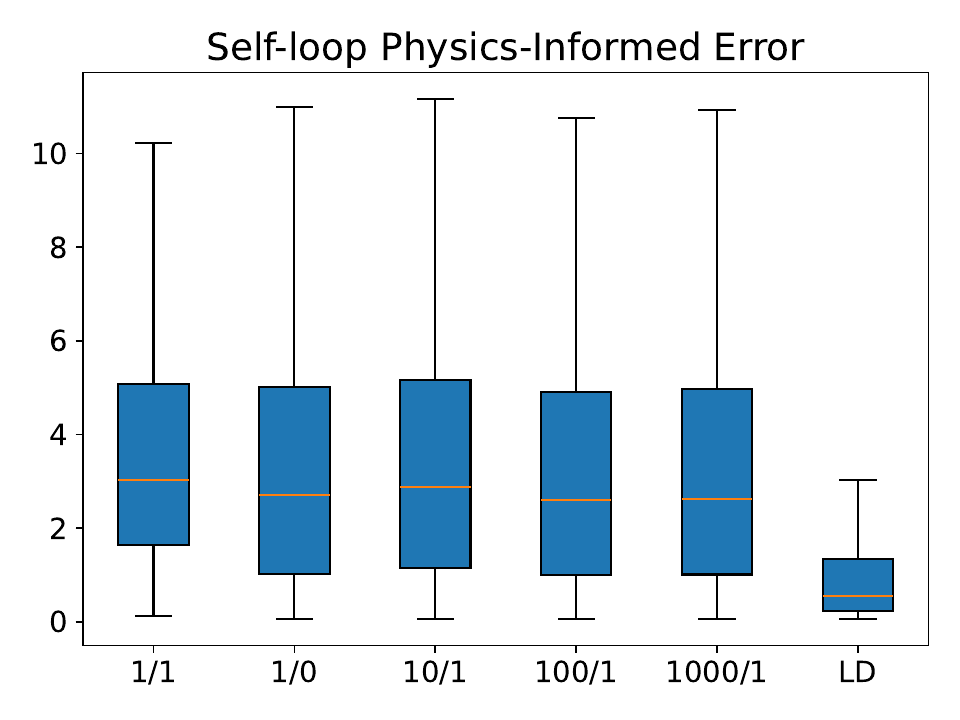}
        \captionsetup{font=footnotesize, justification=centering}
    \end{subfigure}
    \caption{Distribution of the results obtained during the experiments to evaluate the effect of different relative weights $\alpha$ and $\beta$ in the loss. We show the results for various ratio of fixed weights ($\alpha / \beta$) and for the Lagrangian dual approach LD.}
    \label{fig:weights}
\end{figure}
Moreover, as shown in Figure \ref{fig:weights}, we found that the use of the Lagrangian dual method led to improved results with respect to the use of fixed weights. Therefore, the following experiments have been performed using an input composed of the concatenation of state, \(I_{s}\) and \(I_{rw}\) from the training set, and the Lagrangian dual method to change the relative weight of the data-driven and physics-informed losses during training, starting from the same initial weight and with the constraint that the data-driven loss is assigned a higher weight, as stated in Section \ref{methodology}.

Subsequently, we proceeded to evaluate the performances of the models trained only with the data-driven approaches and those trained with both losses, along with the introduction of the two implementations of the self-attention mechanism introduced in Section \ref{methodology}. In Table \ref{tab:dynamic_learn_exp0} are shown the results obtained by these experiments. In particular, the table shows for each experiment the average of the MRE and of the physics error, both for the next step prediction and when using the model in self-loop mode for 10 time steps, i.e. when using the current output of the model as state for the input of the following inference. The experiments labeled with a DD are those in which the model has been trained using only the data-driven loss, while those with LD used the Lagrangian dual method.
\begin{table}
    \caption{Results obtained during the Dynamics Learning Experiments with the Real NVP models. Experiments with DD are performed using only the data-driven loss, those with LD used a Lagrangian dual approach. In bold are shown the models that will be further analyzed in the MPC and robustness-to-noise experiments.}
    \centering
    \begin{tabular}{@{}ccccc@{}}
         \toprule
         \textbf{Experiment} & \textbf{MRE} & \begin{tabular}{@{}c@{}}\textbf{Physics} \\ \textbf{Error}\end{tabular} & \begin{tabular}{@{}c@{}}\textbf{MRE} \\ \textbf{self-loop}\end{tabular} & \begin{tabular}{@{}c@{}}\textbf{Physics Error} \\ \textbf{self-loop}\end{tabular} \\
         \midrule\midrule
         RealNVP-DD & 0.33\% & \underline{0.25} & 28.84 & 91.68 \\
         NVPSA1-DD & 0.35\% & \underline{0.25} & 3.07 & 6.65 \\
         \textbf{NVPSA2-DD} & \textbf{0.32\%} & \textbf{\underline{0.25}} & \textbf{2.40} & \textbf{5.83} \\
         RealNVP-LD & 0.36\% & 0.19 & 2.82 & 0.84 \\
         NVPSA1-LD & 0.32\% & \underline{0.18} & 2.15 & 0.40 \\
         \textbf{NVPSA2-LD} & \textbf{\underline{0.32\%}} & \textbf{0.16} & \textbf{1.75} & \textbf{0.31} \\
         \bottomrule
    \end{tabular}
    \label{tab:dynamic_learn_exp0}
\end{table}
The results show that the self-attention mechanism that, in addition to the inertia matrices, used as input for the key and query networks the commanded torque, rather than the output of the Real NVP network, \textit{NVPSA2}, led to the best performances both when using only the data-driven loss and when using both losses. Moreover, the experiments revealed that the model trained with both losses achieved consistently better results both in terms of physics error and in terms of MRE.

To better evaluate the results and confirm their significance, in addition to the average, we analyzed the standard deviation and computed the p-values with the Wilcoxon test. The standard deviation analysis showed that the models trained only with the data-driven loss had a standard deviation higher of an order of magnitude, and that the same difference held between models trained with the same loss with and without self-attention mechanism. When considering the significance test, a p-value is defined as not significant when it is higher than 0.05. We obtained not significant results only in the single-step evaluation. In particular, the differences between the Physics Error of the data-driven models and of RealNVP-LD and NVPSA2-LD were not significant, nor was the difference between the MRE of the NVPSA1-LD and NVPSA2-LD. We underlined these results in Table \ref{tab:dynamic_learn_exp0}. It should be noted that the MRE of NVPSA2-LD is not significant only with respect to the result found for NVPSA1-LD. We argue that this is due to the small changes in angular velocity in a single step, as the p-values obtained when comparing the predictions over 10 time steps are all close 0.

\subsection{MPC Experiments}
\begin{figure}[h!]
    \centering
    \begin{subfigure}[t]{0.49\textwidth}
        \centering
        \includegraphics[width=\columnwidth]{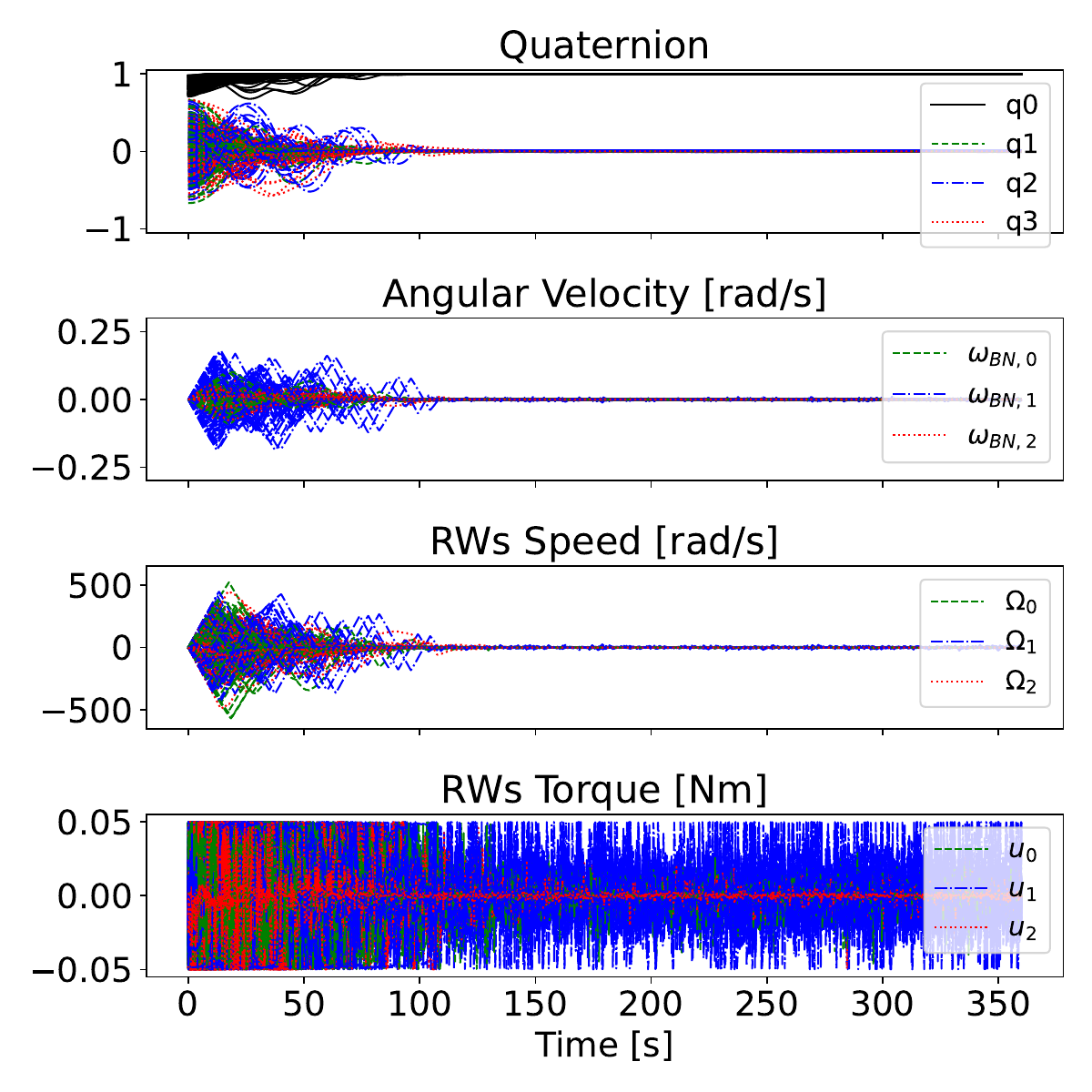}
        \captionsetup{font=footnotesize, justification=centering}
        \caption{NVPSA2-DD}
    \end{subfigure}
    \begin{subfigure}[t]{0.49\textwidth}
        \centering
        \includegraphics[width=\columnwidth]{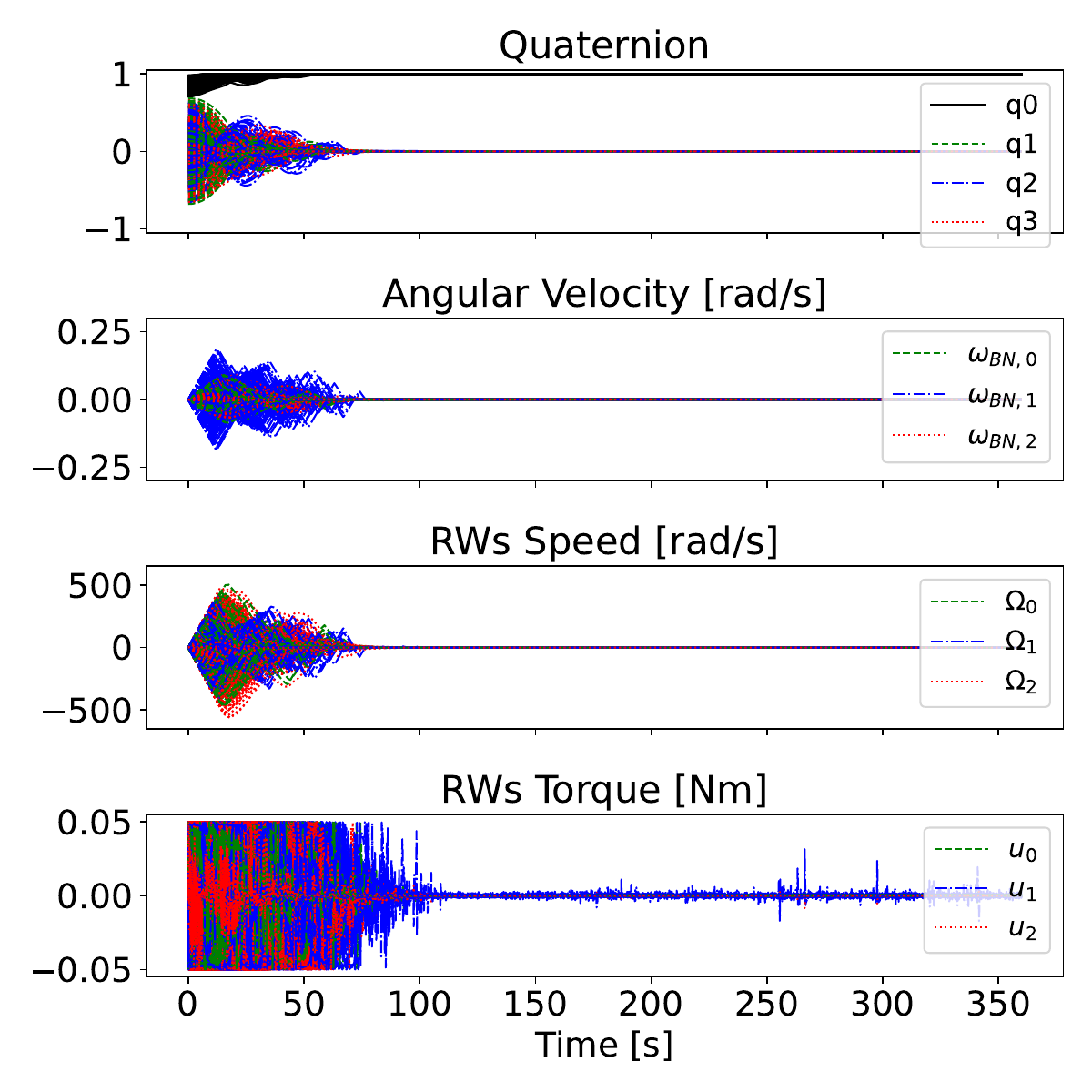}
        \captionsetup{font=footnotesize, justification=centering}
        \caption{NVPSA2-LD}
    \end{subfigure}
    \begin{subfigure}[t]{0.49\textwidth}
        \centering
        \includegraphics[width=\columnwidth]{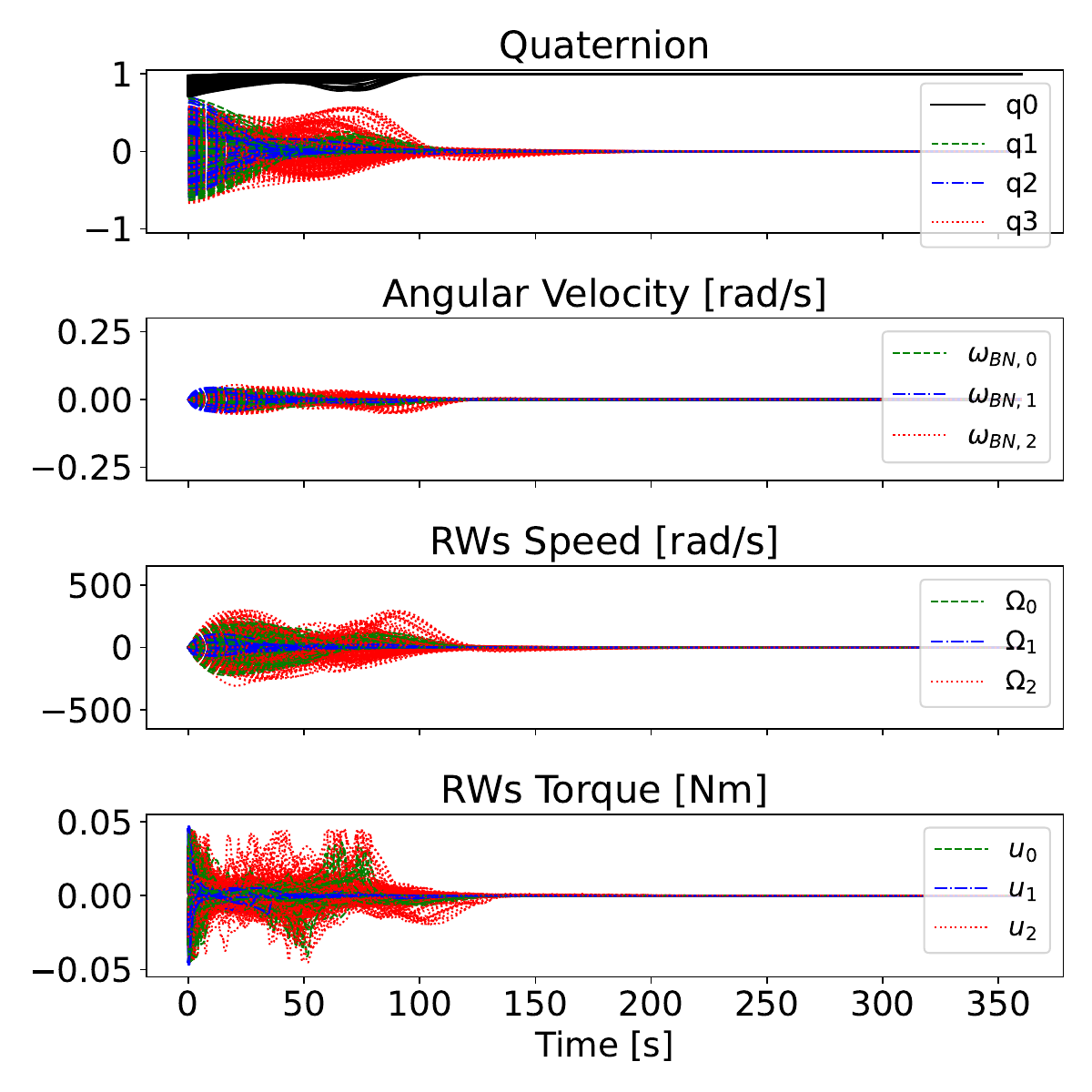}
        \captionsetup{font=footnotesize, justification=centering}
        \caption{Linear MPC}
    \end{subfigure}
    \begin{subfigure}[t]{0.49\textwidth}
        \centering
        \includegraphics[width=\columnwidth]{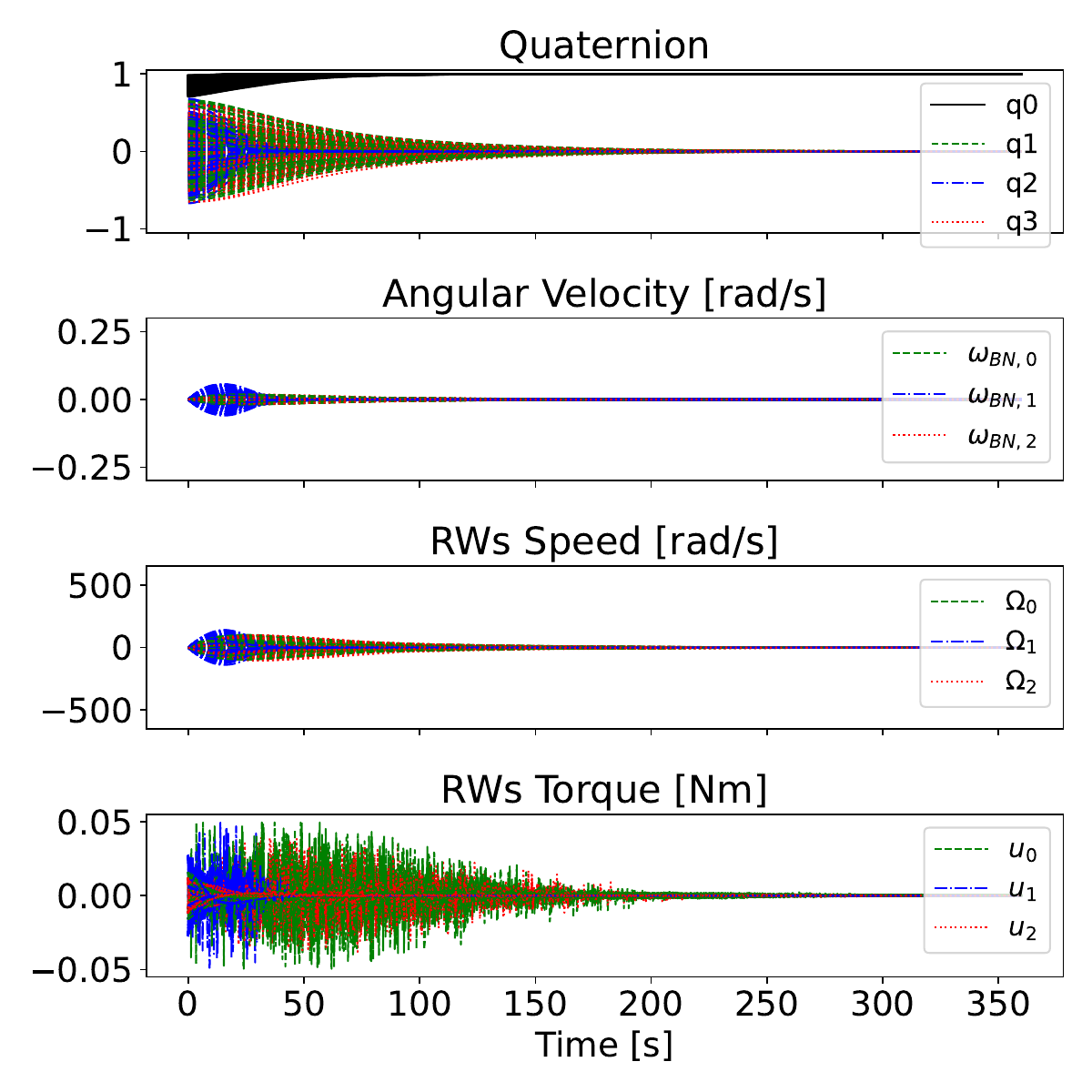}
        \captionsetup{font=footnotesize, justification=centering}
        \caption{Non-Linear MPC}
    \end{subfigure}
    \caption{300 MC simulation with parameter estimation errors, state estimation noise and RWs friction for NVPSA2-DD (upper left), NVPSA2-LD (upper right), traditional MPC with linear (lower left) and non-linear (lower right) dynamics.}
    \label{fig:traj_MPC_noise_est}
\end{figure}

\begin{figure}[!h]
    \centering
    \begin{subfigure}[t]{0.47\textwidth}
        \centering
        \includegraphics[width=1\textwidth]{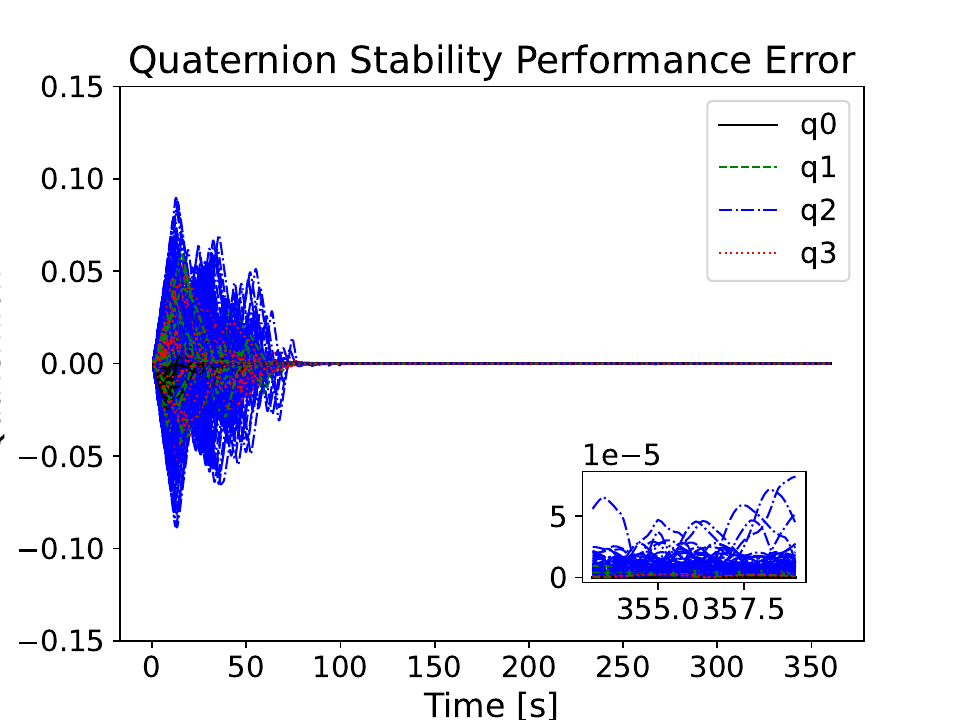}
        \captionsetup{font=footnotesize, justification=centering}
    \end{subfigure}
    \begin{subfigure}[t]{0.47\textwidth}
        \centering
        \includegraphics[width=1\textwidth]{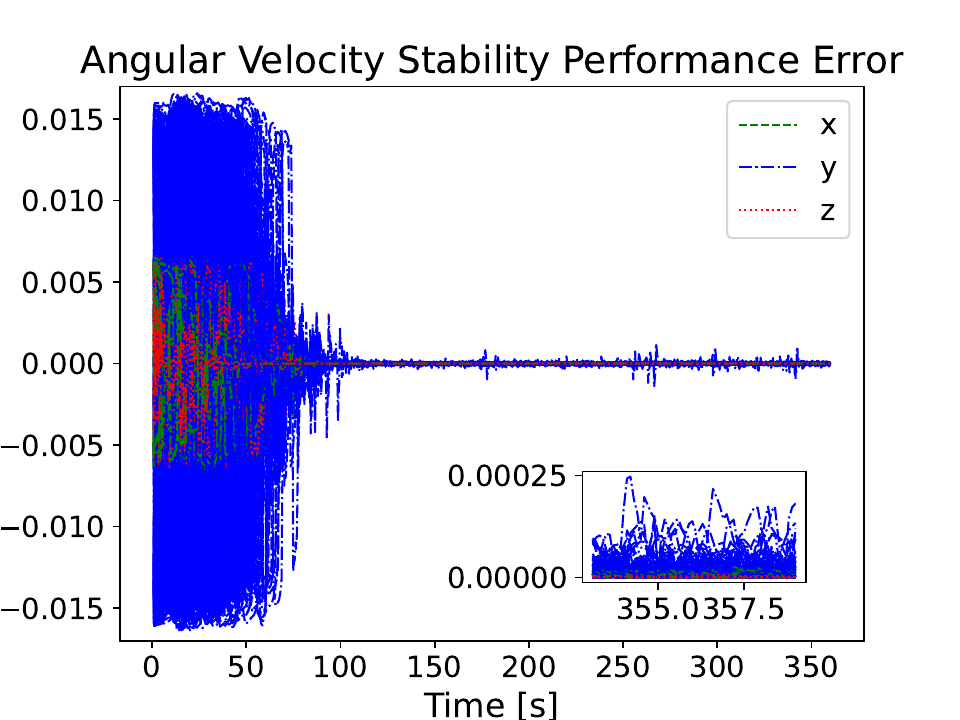}
        \captionsetup{font=footnotesize, justification=centering}
    \end{subfigure}
    \begin{subfigure}[t]{0.47\textwidth}
        \centering
        \includegraphics[width=1\textwidth]{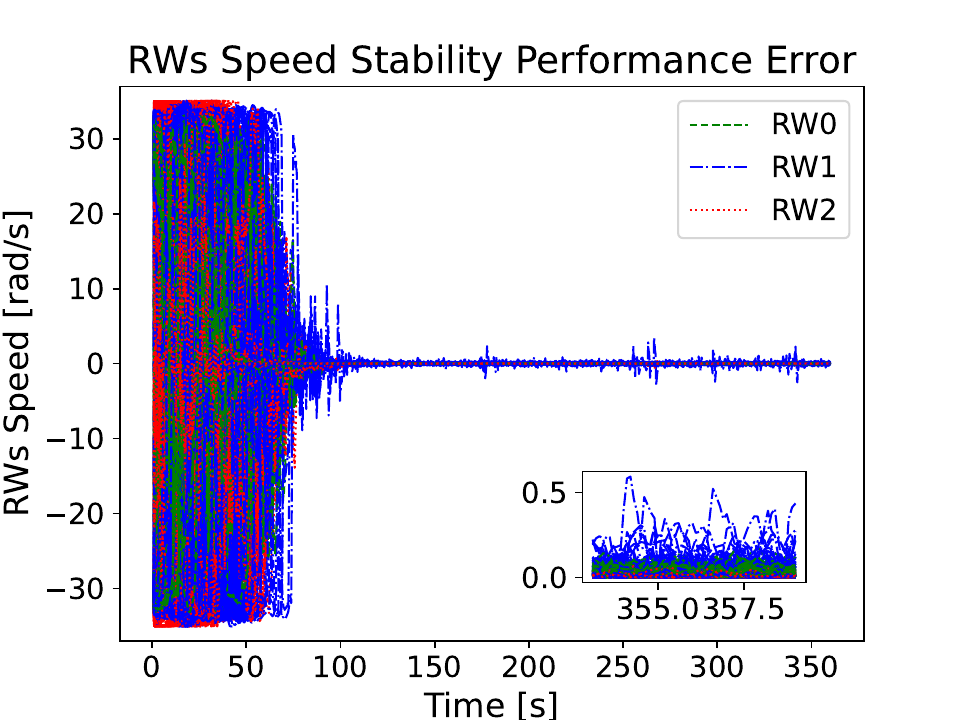}
        \captionsetup{font=footnotesize, justification=centering}
    \end{subfigure}
    \begin{subfigure}[t]{0.47\textwidth}
        \centering
        \includegraphics[width=1\textwidth]{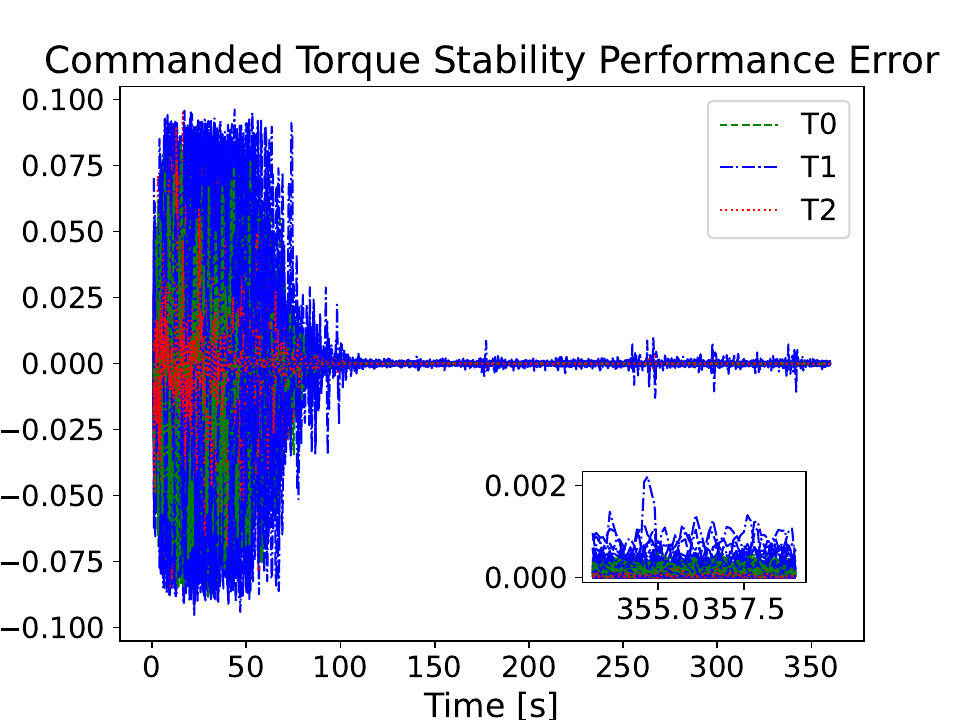}
        \captionsetup{font=footnotesize, justification=centering}
    \end{subfigure}
    \caption{Stability Performance Error for the MPC with NVPSA2-LD under parameter estimation inaccuracies, RWs friction and state estimation noise.}
    \label{fig:spe_ld}
\end{figure}

Given the results presented above, we used the models NVPSA2-LD and NVPSA2-DD as dynamics predictor in a MPC controller to compare the performances of the best performing model trained with and without the physics-informed loss. Comparing them with the two traditional MPC controllers introduced in Section \ref{sec:trad_mpc}. In Figure \ref{fig:traj_MPC_noise_est} we present the trajectories of 300 Monte Carlo simulations when performing rest-to-rest maneuvers over 360 seconds with the model trained only with the data-driven loss (NVPSA2-DD), both the data-driven and the physics-informed loss (NVPSA2-LD), and the two traditional MPC controllers. In these simulations we randomly changed the orbital position, the initial satellite attitude, as well as introduce parameter estimation errors for the satellite weight (up to 20\%) and for the satellite inertia matrix (up to 10\%). Moreover, state estimation was subject to a Guassian random noise of up to 3\% and the reaction wheels were subject to friction, as described in Section \ref{sec:eval_metrics_exp}.
\begin{table}[!htb]
    \caption{Results of the experiments performed to study the robustness-to-noise of the models when used in a MPC with several sources of noise. We show the median over 300 MC simulations.}
    \centering
    \begin{tabular}{@{}ccc@{}}
         \toprule
         \textbf{Experiment} & \begin{tabular}{@{}c@{}}\textbf{Steady-State Error} \\ \textbf{[degrees]}\end{tabular} & \textbf{Settling Time [s]} \\
         \midrule\midrule
         NVPSA2-DD & 0.5865 & 65.0\\
         NVPSA2-LD & 0.0464 & \textbf{42.8}\\
         Non-Linear MPC & 0.1215 & 186.2\\
         Linear MPC & \textbf{0.0024} & 114.1\\
         \bottomrule
    \end{tabular}
    \label{tab:noise_res}
\end{table}
The results show that all models are able to reach the required state with a steady-state error below 1 degree, but the model trained only with the data-driven loss is less stable. Additionally, it can be seen that the MPC with the physics-informed model (NVPSA2-LD) is able to reach a good convergence error, given the noise present in the environment, in the minimum amount of time among all controller tested. Though, it tends to produce higher torques even in steady-state. In Table \ref{tab:noise_res} we show the median settling time, computed as the time required to reach a pointing error below 1 degree, and the median of the steady-state error over the last minute of the trajectory, i.e. the angle in degrees between the current and target quaternion. The results show that the MPC with state space model requires considerably more time to reach convergence, while being able to reach a more accurate attitude, with respect to the MPC with NVPSA2-LD, which is the fastest, i.e. 114 seconds for the state space model against the 43 seconds required by NVPSA2-LD. Moreover, the median settling time for the traditional non-linear MPC is approximately 3 minutes. Finally, Figure \ref{fig:spe_ld} shows the stability performance error for the quaternion, the angular velocity of the satellite, the velocities of the RWs and torque applied to the RWs when using the physics-informed model NVPSA2-LD.

\subsection{Edge Inference} \label{inf_time}
In Table \ref{tab:inference_latency} we show the mean and standard deviation of the inference latency for the three neural network architectures in 3 edge devices: Jetson Nano, Raspberry PI 4B and AMD GX-412HC. In the Jetson Nano we tested 4 different combinations of maximum power consumption and target hardware, i.e. CPU or GPU. Given the small size of the model, the results suggest that it is better to perform inference only in the CPU, because the overhead associated to the data movement to the GPU is higher than the improvement in performance. When considering only the CPUs it can be seen that the bigger model (NVPSA2) could be used with a frequency between 100Hz and 380Hz.
\begin{table}
    \centering
    \caption{Inference latency tests on different edge devices and architectures. For each device and architecture, we show the average and standard deviation of the latency, in milliseconds, over 1000 inferences.}
    \begin{tabular}{@{}cccc@{}}
        \toprule
        Device & RealNVP[ms] & NVPSA1[ms] & NVPSA2[ms] \\
        \midrule
        \midrule
        AMD GX-412HC & 2.30$\pm$0.63 & 2.69$\pm$0.68 & 2.62$\pm$0.09 \\
        Raspberry PI 4B & 7.91$\pm$1.55 & 8.92$\pm$1.59 & 9.12$\pm$1.23 \\
        Jetson Nano CPU 5W & 5.07$\pm$0.28 & 6.13$\pm$0.30 & 6.19$\pm$0.45 \\
        Jetson Nano CPU 10W & 3.26$\pm$1.62 & 3.93$\pm$0.19 & 3.98$\pm$0.35 \\
        Jetson Nano GPU 5W & 15.21$\pm$58.26 & 19.44$\pm$57.74 & 18.86$\pm$61.17 \\
        Jetson Nano GPU 10W & 9.19$\pm$35.77 & 10.65$\pm$37.76 & 11.84$\pm$71.73 \\
        \bottomrule
    \end{tabular}
    \label{tab:inference_latency}
\end{table}

\section{CONCLUSIONS}\label{conclusion}
In this work, we introduced a novel framework for learning spacecraft attitude dynamics using a Real NVP neural network augmented with a self-attention mechanism and trained via a hybrid loss that combines data-driven supervision and physics-informed penalty. The relative weight of the two loss terms has been dynamically changed during training via the Lagrangian dual approach. Our analysis demonstrated that incorporating physical knowledge in the training as an inductive bias significantly enhances model performance, with improvements in terms of Mean Relative Error (MRE) ranging from 27.08\% to 90.22\% when forecasting 10 time steps. Embedding the learned models into an MPC framework further validated their practical utility, showing robust closed-loop control performance and resilience to observational noise and RWs friction. This evaluation was carried out by showing the trajectory distributions of 300 Monte Carlo simulations and by evaluating their average steady-state errors, along with the settling time, for both purely data-driven and physics-informed models, and comparing their performances with traditional MPC solutions. Achieving an improvement in terms of settling time of about 62\% compared to the best performing traditional MPC. Additionally, the Stability Performance Error (SPE) was computed to evaluate the stability of the proposed controller. Across all metrics, the physics-informed model consistently achieved the best results compared to the data-driven approach, demonstrating superior robustness. Although the best results were achieved with 10-step predictions, we believe that larger networks and longer horizons could yield even greater accuracy, though at increased computational cost. We believe that the proposed framework has promising generalization capabilities and could be used to address complex non-linear dynamics in future works, such as those involved in satellite berthing or docking with non-cooperative targets, or for the control of under-actuated spacecraft. These scenarios present greater modeling and control challenges due to factors like discontinuous dynamics, partial observability, and limited actuation, which align well with the strengths of combined data-driven and physics-informed learning. Finally, we envision the integration of traditional and AI-based approaches to leverage the strengths of both, enabling faster and more accurate attitude control, even under non-nominal conditions. In the context of this work, a potential robust solution may see the integration of the MPC with physics-informed dynamics for the highly dynamic maneuvers and steady-state dynamics when close to the target attitude.

\section*{ACKNOWLEDGMENT}
This work has been developed with the contribution of the Politecnico di Torino Interdepartmental Centre for Service Robotics (PIC4SeR) (https:// pic4ser.polito.it) and Argotec (https://argotecgroup.com). Thanks to Manuel Pecorilla from Argotec who helped with valuable suggestions. Funding: This work was supported by the the project PNRR-NGEU which has received funding from the MUR – DM 117/2023.

\section*{Declaration of generative AI and AI-assisted technologies in the writing process}
During the preparation of this work the author(s) used Gemini and Copilot in order to avoid syntactical errors and improve readability. After using this tool/service, the author(s) reviewed and edited the content as needed and take(s) full responsibility for the content of the publication.

%% The Appendices part is started with the command \appendix;
%% appendix sections are then done as normal sections
%\appendix
%\section{Example Appendix Section}
%\label{app1}

\section*{CRediT authorship contribution statement} 
\textbf{Carlo Cena}: Conceptualization, Investigation, Software, Writing. \textbf{Mauro Martini}: Conceptualization, Writing. \textbf{Marcello Chiaberge}: Funding acquisition, Project administration.

\bibliographystyle{elsarticle-harv}
\bibliography{refs}

\end{document}